\documentclass[11pt]{article}
\usepackage{authblk}
\usepackage{eacl2017}
\usepackage{times}
\usepackage{url}
\usepackage{latexsym}

\eaclfinalcopy 
\def\eaclpaperid{449} 

\usepackage{mathrsfs}
\usepackage{verbatim}
\usepackage{times}
\usepackage{url}
\usepackage{latexsym}
\usepackage{color}
\usepackage{float}
\usepackage{array}
\usepackage{amsmath,amsfonts,amssymb}
\usepackage{algorithm}
\usepackage{algpseudocode}
\usepackage{graphicx}
\usepackage{subcaption}
\usepackage{booktabs}
\usepackage{multirow}
\usepackage[inline]{enumitem}
\usepackage{pifont}
\usepackage{fancyvrb}
\usepackage{footnote}

\makesavenoteenv{tabular}
\makesavenoteenv{table}

\usepackage[font=small]{caption}

\newcommand{\dataname}{\textsc{Draw-1k}}

\newcommand{\slots}{\mathcal{C}(T)}
\newcommand{\numbers}{\mathcal{Q}(\bx)}
\newcommand{\dspace}[1]{\mathcal{D}({#1})}

\usepackage{stmaryrd}

\def\by{{\boldsymbol{y}}}
\def\bz{{\boldsymbol{z}}}
\def\bx{{\boldsymbol{x}}}

\usepackage{xargs}
\usepackage[colorinlistoftodos,prependcaption,textsize=tiny]{todonotes}

\newcommand{\ignore}[1]
{
}

\newcommand*{\Let}[2]{\State {#1} $\gets$ {#2}}
\title{Annotating Derivations: A New Evaluation Strategy and Dataset for Algebra Word Problems}

\author[1]{\bf Shyam Upadhyay}
\author[2]{\bf Ming-Wei Chang}
\affil[1]{University of Illinois at Urbana-Champaign, IL, USA}
\affil[2]{Microsoft Research, Redmond, WA, USA}
\affil[ ]{\tt  upadhya3@illinois.edu}
\affil[ ]{\tt  minchang@microsoft.com}

\date{}

\begin{document}
\maketitle

\begin{abstract}
  We propose a new evaluation for automatic solvers for algebra word
problems, which can identify mistakes that existing
evaluations overlook. Our proposal is to evaluate such solvers using {\em derivations}, which reflect how an equation system was constructed from the word problem.  To accomplish
this, we develop an algorithm for checking the equivalence between
two derivations, and show how derivation annotations can be
semi-automatically added to existing datasets. To make our experiments
more comprehensive, we include the derivation annotation for \dataname, a new dataset containing
1000 general algebra word problems. In our experiments, we found that the annotated derivations enable a more accurate evaluation of automatic solvers than
previously used metrics.
We release derivation annotations for over 2300 algebra word problems for future evaluations. 


\end{abstract}

\section{Introduction}
\label{sec:introduction}
\begin{figure}[t]
  \centering
  \includegraphics[scale=0.45]{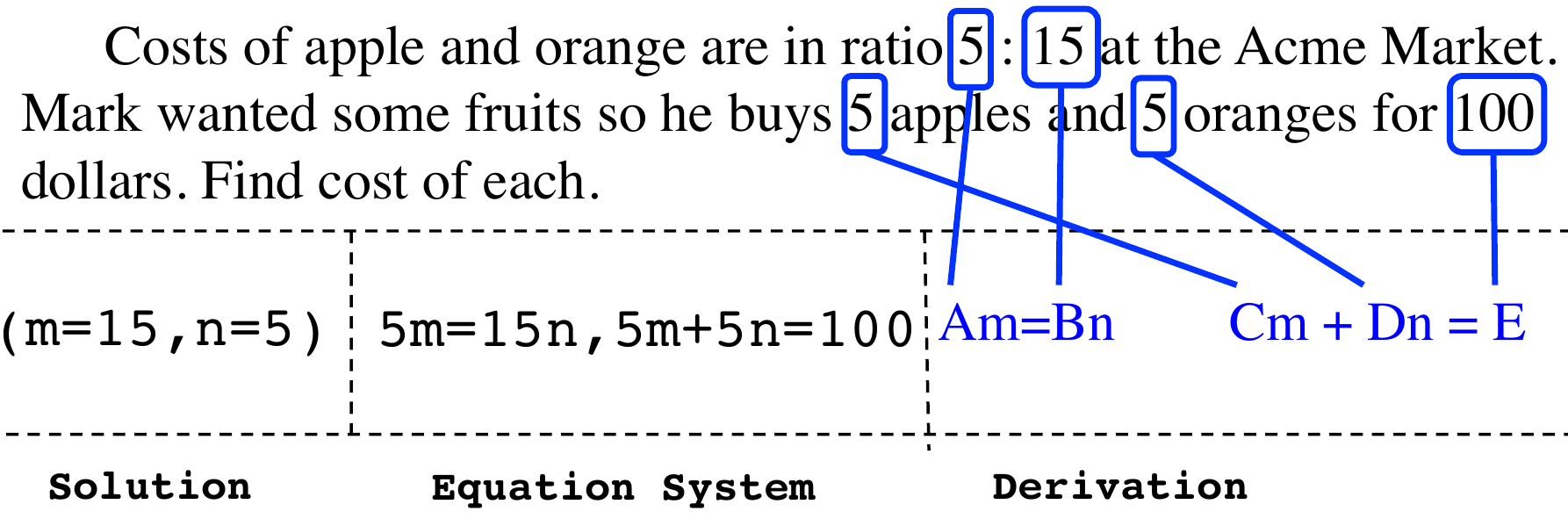}
  \caption{\footnotesize An algebra word problem with its {\em solution},
    {\em equation system} and {\em derivation}. Evaluating solvers on
    derivation is more reliable than evaluating on solution or
    equation system, as it reveals errors that other metric overlook.}
  \label{fig:firstex}
\end{figure}

Automatically solving math reasoning problems is a long-pursued goal
of AI~\cite{newell1959report,Bobrow1964}.  Recent
work~\cite{Kushman14,MSRA2015,TACL692} has focused on developing
solvers for {\em algebra word problems}, such as the one shown in
Figure~\ref{fig:firstex}. Developing a solver for word problems can
open several new avenues, especially for online education and
intelligent tutoring systems~\cite{Kang2016}. In addition, as solving
word problems requires the ability to understand
and analyze 
natural language,
it serves as a good test-bed for evaluating progress towards goals of artificial
intelligence~\cite{clark2016my}.

An automatic solver finds the solution of a given word problem by
constructing a {\em derivation}, consisting of an un-grounded equation
system\footnote{Also referred to as a {\em template}. We use these two
  terms interchangeably.} (${\tt\{Am=Bn,Cm+Dn=E\}}$ in
Figure~\ref{fig:firstex}) and {\em alignments} of numbers in the text
to its coefficients (blue edges). The derivation identifies a grounded
equation system ${\tt\{5m=15n,5m+5n=100\}}$, whose {\em solution} can
then be generated to answer the problem. A derivation precisely describes {\em how} the
grounded equation system was constructed from the word problem by the
automatic solver. On the other hand, the grounded equation systems and
the solutions are less informative, as they do {not} explain which
span of text aligns to the coefficients in the equations.

While the derivation is clearly the most informative structure,
surprisingly, {no} prior work evaluates automatic solvers using
derivations directly.  To the best of our knowledge, none of the current datasets contain
human-annotated derivations, possibly due to the belief that the
current evaluation metrics are sufficient and the benefit of
evaluating on derivations is minor.  Currently, the most popular
evaluation strategy is to use {\em solution accuracy} \cite{Kushman14,Verb14,MSRA2015,TACL692,Baidu2015:EMNLP,huang-EtAl:2016:P16-12}, which computes whether the
solution was correct or not,
as this is an easy-to-implement metric.  Another evaluation strategy was proposed
in~\cite{Kushman14}, which finds an approximate derivation from the
gold equation system and uses it to compare against a predicted
derivation. We follow~\cite{Kushman14} and call this evaluation strategy the
{\em equation accuracy}.~\footnote{Note that an approximation of the derivation is
  necessary, as there is no annotated derivation. From the brief
  description in their paper and the code released by~\newcite{Kushman14}, we found
  that their implementation {assumes} that the first derivation that
  matches the equations and generates the correct solution is the
  correct reference derivation against which predicted derivations are
  then evaluated. }

In this work, we argue that evaluating solvers against human labeled
derivation is important. Existing evaluation metrics, like solution
accuracy are often quite generous --- for example, an incorrect
equation system, such as,
\begin{equation}
\begin{array}{cc}
\{m+5=n+15, & m+n=15+5 \},
\end{array}
\end{equation}
can generate the correct solution of the word problem in
Figure~\ref{fig:firstex}. 
While equation accuracy appears to be a stricter metric than solution accuracy, our experiments show that the approximation can mislead evaluation, by assigning higher scores to an inferior solver. Indeed, a correct equation system,
($5m=15n,5m+5n=100$), can be generated by using a wrong template, ${\tt A}m={\tt B}n,
{\tt A}m + {\tt A}n = {\tt C}$, and aligning numbers in the text to coefficients incorrectly. 
We show that without knowing the correct derivation at evaluation time, a solver can be awarded for the wrong reasons. 

The lack of annotated derivations for word problems and no clear
definition for comparing derivations present technical difficulties in
using derivation for evaluation.  In this paper, we address these
difficulties and for the first time propose to evaluate the solvers
using {\em derivation accuracy}. To summarize, the contributions of
this paper are:
\begin{itemize}
\item We point out that evaluating using derivations is more precise
  compared to existing metrics. Moreover, contrary to popular belief,
  there is a meaningful gap between the derivation accuracy and existing metrics, as it can discover crucial errors not captured previously.
\item We formally define when two derivations are equivalent, and develop an algorithm that can determine the same. The algorithm is simple to implement, and can
  accurately detect the equivalence even if two derivations have very
  different syntactic forms.
\item We annotated over 2300 word algebra problems\footnote{available at \url{https://aka.ms/datadraw}}
  with detailed derivation annotations, providing high quality labeled
  semantic parses for evaluating word problems.
\end{itemize}


\section{Evaluating Derivations}
\label{sec:eval-deriv}
\begin{table}[t]
    \centering
    \small
    \begin{tabular}{p{2.5cm}cp{3.2cm}}
      \toprule
      Word Problem & $\bx$ & We are mixing a solution of 32\% sodium and another solution of 12\% sodium. How many liters of 32\% and 12\% solution will produce 50 liters of a 20\% sodium solution?
\\ 
      Textual Numbers & $\numbers$ & $\{\text{$32_1, 12_1, 32_2, 12_2, 50, 20$}\}$\\ 
      Equation System  & $\by$ & $32m + 12n=20*50$, $m + n = 50$ \\  
      Solution & & $m = 20$, $n = 30$ \\
      \midrule
      {\bf Template} & $T$ & ${\tt Am+Bn=C*D}$, ${\tt m + n = C}$ \\ 
      {\bf Coefficients} & $\slots$ & ${\tt A,B,C,D}$ \\ 
      {\bf Alignments} & $A$ & $\{\text{$32_1$} \rightarrow {\tt A}$, $\text{$12_1$} \rightarrow {\tt B}$, $\text{50} \rightarrow {\tt C}$, $\text{20} \rightarrow {\tt D}\}$\\
      {\bf EquivTNum} & & \{[$32_1,32_2$], [$12_1,12_2$]\} \\
      {\bf Derivation} & $\bz$ & $(T,A)$\\
      \bottomrule
    \end{tabular}
  \caption{The symbols we used in the paper. Our proposed annotations
    are shown in {\bf bold}. 
Equivalent textual numbers, described in EquivTNum, are distinguished with subscripts.
  }
  \label{tab:notations}
\end{table}



We describe our notation and revisit the notion of derivation introduced in~\cite{Kushman14}. We then formalize the notion of derivation equivalence and provide an algorithm to determine it.
\paragraph{Structure of Derivation}
The word problem in Table~\ref{tab:notations} shows our
notation, where our proposed annotations are shown in {\bf bold}.
%
We denote a word problem by $\bx$ and an {\em equation system} by
$\by$. 

An un-grounded equation system (or {\em template}) $T$ is a family of equation systems parameterized by a set of coefficients $\slots = \{c_i\}_{i=1}^k$, where each coefficient $c_i$ aligns to a
{\em textual number} (e.g., \emph{four}) in the word problem. We also refer to the coefficients as {\em slots} of the template. We use ${\tt (A,B,C,\ldots)}$ to represents coefficients and ${\tt
  (m,n,\ldots)}$ to represent the unknown variables in the
templates. 

Let $\numbers$ be the set of all the {textual numbers} in the problem $\bx$,
and $\slots$ be the coefficients to be determined in the template $T$.
An {\em alignment} is a set of tuples $A=\{(q,c) \mid q \in \numbers,c
\in \slots \cup\{\epsilon\}\}$ aligning textual numbers to coefficient slots, where a tuple $(q,\epsilon)$
indicates that the number $q$ is not relevant to the final equation
system. 

Note that there may be multiple semantically equivalent textual numbers. e.g., in Figure~\ref{fig:firstex}, either of the 32 can
be aligned to coefficient slot {\tt A} in the template. These equivalent textual numbers are marked 
in the EquivTNum field in the annotation. If two textual numbers $q,q'\in$ EquivTNum, then we can align a coefficient slot to either $q$ or $q'$, and generate a equivalent alignment.

An alignment $A$ and a template $T$ together identify a {\em
  derivation} $\bz=(T,A)$ of an equation system. Note that there may
be multiple valid derivations, using one of the equivalent alignments.
We assume there exists a routine \texttt{Solve}($\by$) that
find the solution of an equation system.  We use a Gaussian
elimination solver for our \texttt{Solve} routine.
We use hand-written rules and the quantity normalizer in Stanford CoreNLP~\cite{manning2014} to identify textual numbers.

\paragraph{Derivation Equivalence}
\begin{algorithm}[t]
  \footnotesize
  \caption{Evaluating Derivation}
  \label{algo:eval}
  \begin{algorithmic}[1]
    \Require{Predicted $(T_p,A_p)$ and gold $(T_g,A_g)$ derivation 
    }
    \Ensure{1 if predicted derivation is correct, 0 otherwise}
    \If{$|\mathcal{C}(T_p)| \neq |\mathcal{C}(T_g)|$}
    \Comment{different \# of coeff. slots}
    \State{return 0}
    \EndIf
    \Let{$\Gamma$}{\Call{TemplEquiv}{$T_p$,$T_g$}} \label{lst:ingamma}
    \If{$\Gamma = \emptyset$}
    \Comment{not equivalent templates}
    \State{return 0}
    \EndIf
    \If{\Call{AlignEquiv}{$\Gamma,A_p,A_g$}} \Comment{Check alignments}
    \State{return 1}
    \EndIf
    \State{return 0} 
    \\\hrulefill
    \Procedure{TemplEquiv}{$T_1,T_2$} \label{lst:procedure}
    \State{}
    \Comment{Note that here $|\mathcal{C}(T_1)| = |\mathcal{C}(T_2)|$ holds}
    \State {$\Gamma \leftarrow \emptyset$}
    \For{each 1-to-1 mapping $\gamma$ : $\mathcal{C}(T_1)
      \rightarrow \mathcal{C}(T_2)$} \label{lst:gamma}
    \Let{match}{True}
    \For{$t=1 \cdots R$}
    \Comment{$R$ : Rounds}
    \State{Generate random vector $\mathbf{v}$} \label{lst:random}
    \State{$A_1\leftarrow\{(\mathbf{v}_i\rightarrow
      c_i)\}$,$A_2\leftarrow\{(\mathbf{v}_i\rightarrow
      \gamma(c_i))\}$}
    \If{{\tt Solve}($T_1,A_1$) $\neq$ {\tt Solve}($T_2,A_2$)}
    \Let{match}{False}; {\bf break}
    \EndIf
    \EndFor
    \State{{\bf if} match {\bf then} $\Gamma \leftarrow \Gamma \cup \{\gamma\}$}
    \EndFor
    \State{return {$\Gamma$}}\Comment{$\Gamma \neq \emptyset$ iff the
      templates are equivalent} \label{lst:retproc}
    \EndProcedure
    \\\hrulefill
    \Procedure{AlignEquiv}{$\Gamma,A_1,A_2$}
    \For{mapping $\gamma \in \Gamma$}
    \State{{{\bf if} following holds true},
      $$
      (q,c)\in A_1\iff \{ (q,\gamma(c)) \text{ or }(q',\gamma(c)) \}\in A_2  
      $$}
    \State{where $(q',q) \in$ EquivTNum} \label{lst:alignmatch}
    \State{{\bf then} return 1}
    \State{{\bf end if}}
    \EndFor
    \State{return 0}
    \EndProcedure
  \end{algorithmic}
\end{algorithm}

We define two derivations ($T_1,A_1$) and ($T_2,A_2$) to be equivalent {\em iff}
the corresponding templates $T_1,T_2$ and alignments $A_1,A_2$ are
equivalent.

Intuitively, two templates $T_1,T_2$ are equivalent if they can
generate the same space of equation systems -- i.e., for every
assignment of values to slots of $T_1$, there exists an assignment of
values to slots of $T_2$ such that they generate the same equation
systems. For instance, template (\ref{eq:t1}) and (\ref{eq:t2}) below are equivalent
\begin{equation}
  \begin{array}{cc}
    m = {\tt A} + {\tt B}n & m = {\tt C}-n \label{eq:t1}
  \end{array}
\end{equation}
\begin{equation}
  \begin{array}{cc}
    m + n = {\tt A} & m - {\tt C}n = {\tt B}.\label{eq:t2}
  \end{array}
\end{equation}
because after renaming $({\tt A,B,C})$ to $({\tt B,C,A})$ respectively
in template (\ref{eq:t1}), and algebraic manipulations, it is
identical to template (\ref{eq:t2}). We can see that any assignment of
values to {\em corresponding} slots will result in the same equation
system.

Similarly, two alignments $A_1$ and $A_2$ are equivalent if
corresponding slots from each template align to the same textual
number. For the above example, the alignment \{$1 \rightarrow {\tt A}, 3
\rightarrow {\tt B},4 \rightarrow {\tt C}$\} in template (\ref{eq:t1}), and alignment \{$1
\rightarrow {\tt B}, 3 \rightarrow {\tt C},4 \rightarrow {\tt A}$\} in template  (\ref{eq:t2}) are
equivalent. Note that the alignment \{$1 \rightarrow {\tt A}, 3 \rightarrow
{\tt B},4 \rightarrow {\tt C}$\} for (\ref{eq:t1}) is not equivalent to \{$1
\rightarrow {\tt A}, 3 \rightarrow {\tt B},4 \rightarrow {\tt C}$\} in (\ref{eq:t2}),
because it does not respect variable renaming. Our definition also allows two alignments to be equivalent, if they use textual numbers in equivalent positions for corresponding slots (as described by EquivTNum field). 

In the following, we carefully explain how template and alignment
equivalence are determined algorithmically. Algorithm~\ref{algo:eval}
shows the complete algorithm for comparing two derivations.

\paragraph{Template Equivalence}
We propose an approximate procedure {\sc TemplEquiv}
(line~\ref{lst:procedure}) that detects equivalence between two
templates.  The procedure relies on the fact that under appropriate
renaming of coefficients, two equivalent templates will generate
equations which have the same solutions, for all possible coefficient
assignments.

For two templates $T_1$ and $T_2$, with the same number of
coefficients $|\mathcal{C}(T_1)|=|\mathcal{C}(T_2)|$, we represent a
choice of renaming coefficients by $\gamma$, a 1-to-1 mapping from
$\mathcal{C}(T_1)$ to $\mathcal{C}(T_2)$. The two templates are
equivalent if there exists a $\gamma$ such that solutions of the equations
identified by $T_1$ and $T_2$ are same, for all possible coefficient
assignments.
The {\sc TemplEquiv} procedure exhaustively tries all possible
renaming of coefficients (line~\ref{lst:gamma}), checking if the
solutions of the equation systems generated from a random assignment
(line~\ref{lst:random}) match exactly. It declares equivalence if for
a renaming $\gamma$, the solutions match for $R=10$ such random
assignments.\footnote{Note that this procedure is a Monte-Carlo
  algorithm, and can be made more precise by increasing $R$. We
  found making $R$ larger than 10 did not have an impact on the
  empirical results.}  The procedure returns all renamings $\Gamma$ of
coefficients between two templates under which they are equivalent
(line \ref{lst:retproc}). We discuss its effectiveness in \S\ref{sec:annotating}.

\paragraph{Alignment Equivalence}
The {\sc TemplEquiv} procedure returns every mapping $\gamma$ in
$\Gamma$ under which the templates were equivalent (line
\ref{lst:ingamma}). Recall that $\gamma$ identifies corresponding
slots, $c$ and $\gamma(c)$, in $T_1$ and $T_2$ respectively. We
describe alignment equivalence using these mappings.

Two alignments $A_1$ and $A_2$ are equivalent if corresponding slots
(according to $\gamma$) align to the same textual
number. More formally, if we find a mapping $\gamma$ such that for
each tuple ($q,c$) in $A_1$ there is ($q,\gamma(c)$) in $A_2$, then
the alignments are equivalent (line \ref{lst:alignmatch}). We allow
for equivalent textual numbers (as identified by EquivTNum field) to
match when comparing tuples in alignments.

The proof of correctness of Algorithm \ref{algo:eval} is sketched in the appendix.
Using Algorithm \ref{algo:eval}, we can define {\em derivation accuracy}, to be 1 if the predicted derivation $(T_p,A_p)$ and the reference derivation $(T_g,A_g)$ are equivalent, and 0 otherwise. 

\paragraph{Properties of Derivation Accuracy} By comparing derivations, we can ensure that the following errors are detected by the evaluation.

Firstly, correct solutions found using incorrect equations will be
penalized, as the template used will not be equivalent to reference
template. Secondly, correct equation system obtained by an incorrect
template will also be penalized for the same reason. Lastly, if the
solver uses the correct template to get the correct equation system,
but aligns the wrong number to a slot, the alignment will not be
equivalent to the reference alignment, and the solver will be
penalized too.

We will see some illustrative examples of above errors in \S\ref{sec:case-study}. Note that the currently popular evaluation metric of solution accuracy will not detect {\em any} of these error types.


\section{Annotating Derivations}
\label{sec:annotating}
As none of the existing benchmarks contain derivation annotations, we
decided to augment existing datasets with these annotations.
We also annotated \dataname, a new dataset of 1000 general algebra word
problems to make our study more comprehensive. Below, we describe how we reduced annotation effort by semi-automatic generated
some annotations.

Annotating gold derivations from scratch for all problems is time
consuming. However, not all word problems require manual annotation
-- sometimes all numbers appearing in the equation system can be uniquely
aligned to a textual number without ambiguity. For such problems, the
annotations are generated automatically.\footnote{Annotations for {\em
    all} problems are manually verified later.} We identify word
problems which have at least one {\em alignment ambiguity} -- multiple
textual numbers with the same value, which appears in the equation
system.  A example of such a problem is shown in Figure~\ref{fig:firstex}, where there are three textual numbers with value 5, which appears in the equation system. Statistics for the number of word problems with such ambiguity is shown in
Table~\ref{tab:comp}.  

We only ask annotators to resolve such
alignment ambiguities, instead of annotating the entire derivation. If
more than one alignments are genuinely correct (as in word problem of 
Table~\ref{tab:notations}), we ask the annotators to mark both (using
the {\em EquivTNum} field). This ensures our derivation annotations are
{\em exhaustive} -- all correct derivations are marked. With the correct
alignment annotations, templates for all problems can be easily
induced.

\paragraph{Annotation Effort}
To estimate the effort required to annotate derivations, we timed our annotators when annotating 50 word problems (all involved alignment ambiguities). As a control, we also asked annotators to annotate the entire derivation from scratch (i.e., only provided with the word problem and equations), instead of only fixing alignment ambiguities. When annotating from scratch, annotators took an average of 4 minute per word problem, while when fixing alignment ambiguities this time dropped to average of 1 minute per word problem. 
We attained a inter-annotator agreement of 92\% (raw percentage agreement), with most disagreements arising on EquivTNum field.\footnote{These were adjudicated on by the first author.}

\paragraph{Reconciling Equivalent Templates}
The number of templates has been used as a measure of dataset
diversity~\cite{MSRA2015,huang-EtAl:2016:P16-12}, however prior work
did not reconcile the equivalent templates in the dataset. Indeed, if two templates are equivalent, we can replace one with the other and still generate the correct equations. Therefore, after getting human judgements on alignments, we
reconcile all the templates using {\sc TemplEquiv} as the final step of annotation.

{\sc TemplEquiv} is quite
effective (despite being approximate), reducing the number of templates by at least 20\% for all
datasets (Table~\ref{tab:comp}). We did not find any false positives generated by the {\sc TemplEquiv}
in our manual examination. The reduction in
Table~\ref{tab:comp} clearly indicates that equivalent templates are
quite common in all datasets, and number of templates (and hence, dataset diversity) can be
significantly overestimated without proper reconciliation.


\section{Experimental Setup}
\label{sec:setup}
\begin{table}
  \centering
  \footnotesize
  \begin{tabular}{p{2cm}cccc}
    \toprule
    Dataset & \dataname & {\sc Alg-514} & {\sc Dolphin-L}\\
    \midrule
    \# problems & {1000} & 514 & 832\\  %
    w/ ambiguity & 21\% & 23\% & 35\% \\
    vocab. & {2.21k} & 1.83k & 0.33k\\ %
    
    \midrule
    & \multicolumn{3}{c}{Number of Templates} \\
    \midrule
    before &  329 & 30 & 273 \\
    after & {224} & 24 & 203\\  %
    \% reduction & 32\% & 20\% & 25\% \\
    \bottomrule
  \end{tabular}
  \caption{Statistics of the datasets. 
    At least 20\% of problems in each dataset had alignment ambiguities that required human annotations. The number of templates before and after annotation is also shown (reduction $>20\%$).}
  \label{tab:comp}
\end{table}

We describe the three datasets used in our experiments.  Statistics
comparing the datasets is shown in Table~\ref{tab:comp}. In total,
our experiments involve over 2300 word problems.

\paragraph{Alg-514} The dataset {\sc Alg-514} was introduced in ~\cite{Kushman14}. It
consists of 514 general algebra word problems ranging over a
variety of narrative scenarios (distance-speed, object counting,
simple interest, etc.).

\paragraph{Dolphin-L} {\sc Dolphin-L} is the linear-T2 subset of the
{\sc Dolphin} dataset~\cite{MSRA2015}, which focuses on {\em
  number word problems} -- algebra word problems which
describe mathematical relationships directly in the text. All word problems in the linear-T2 subset of the {\sc Dolphin} dataset can be solved using linear equations.

\paragraph{\dataname} {\em D}ive{\em r}se {\em A}lgebra {\em W}ord (\dataname), consists of 1000 word problems crawled
from \url{algebra.com}. Details on the dataset creation can be found
in the appendix. As {\sc Alg-514} was also crawled from \url{algebra.com}, we ensured that there is little overlap between the datasets.

We randomly split \dataname~into train, development and test splits with 600,
200, 200 problems respectively. We use 5-fold cross validation splits
provided by the authors for {\sc Dolphin-L} and {\sc Alg-514}.

\subsection{Evaluation}
\label{sec:metric}
We compare derivation accuracy against the following evaluation
metrics.
\paragraph{Solution Accuracy} We compute solution accuracy by checking
if each number in the reference solution appears in the generated
solution (disregarding order), following previous work~\cite{Kushman14,MSRA2015}. 

\paragraph{Equation Accuracy}
An approximation of derivation accuracy that is similar to the one
used in \newcite{Kushman14}. We approximate the reference derivation
$\tilde{\bz}$ by randomly chosen from the (several possible)
derivations which lead to the gold $\by$ from $\bx$. Derivation
accuracy is computed against this (possibly incorrect) reference
derivation.  Note that in equation accuracy, the approximation is used instead of
annotated derivation. We include the metric of equation accuracy
in our evaluations to show that human annotated derivation is
necessary, as approximation made by equation accuracy might be problematic.

\subsection{Our Solver}

We train a solver using a simple modeling approach inspired by
\newcite{Kushman14} and \newcite{Baidu2015:EMNLP}.  The solver operates as follows. Given a word
problem, the solver ranks all templates seen during training,
$\Gamma_{train}$, and selects the set of the top-$k$ (we use $k=10$)
templates $\Pi \subset \Gamma_{train}$. Next, all possible derivations $\dspace{\Pi}$ that
use a template from $\Pi$ are generated and scored. The equation system $\hat{\by}$
identified by highest scoring derivation $\hat{\bz}$ is output as the
prediction. Following~\cite{Baidu2015:EMNLP}, we do not model the
alignment of nouns phrases to variables, allowing for tractable
inference when scoring the generated derivations. The solver is trained using a structured
perceptron~\cite{Collins02}.
We extract the following features for a $(\bx,\bz)$ pair,
\paragraph{Template Features.}
Unigrams and bigrams of lemmas and POS tags from the word problem
$\bx$, conjoined with $|\numbers|$ and $|\slots|$.
\paragraph{Alignment Tuple Features.}
For two alignment tuples, $(q_1,c_1),(q_2,c_2)$, we add features
indicating whether $c_1$ and $c_2$ belong to the same equation in the
template or share the same variable. If they belong to the same
sentence, we also add lemmas of the nouns and verbs between $q_1$ and
$q_2$ in $\bx$.
\paragraph{Solution Features.}
Features indicating if the solution of the system identified by the
derivation are integer, negative, non-negative or fractional.


\section{Experiments}
\label{sec:experiments}
\begin{table}
  \centering
  \begin{tabular}{p{1.4cm}cc|c}
    \toprule
    {Setting} & Soln. Acc. & Eqn. Acc. & Deriv. Acc.  \\
    \midrule
    \multicolumn{4}{c}{{\sc Alg-514}}               \\
    \midrule
    TE            & 76.2     & 72.7    & 75.5       \\
    TD            & 78.4     & 73.9    & 77.8       \\
    TD - TE       & 2.2      & 1.2     & {\bf 2.3}  \\
    \midrule
    \multicolumn{4}{c}{\dataname}                   \\
    \midrule
    TE            & 52.0     & 48.0    & 48.0       \\
    TD            & 55.0     & 48.0    & 53.0       \\
    TD - TE       & 3.0      & 0       & {\bf 5.0}  \\
    \midrule
    \multicolumn{4}{c}{{\sc Dolphin}}               \\
    \midrule
    TE            & 55.1     & 50.1    & 44.2       \\
    TD            & 57.5     & 36.8    & 54.9       \\
    TD - TE       & 2.4      & -13.3   & {\bf 10.7} \\
    \bottomrule
  \end{tabular}
  \caption{TE and TD compared using different evaluation metrics. Note
    that while TD is clearly superior to TE due to extra supervision
    using the annotations, only derivation accuracy is able to
    correctly reflect the differences.}
    \label{tab:contributions}
\end{table}
\begin{table}
  \centering
  \begin{tabular}{p{1.45cm}cc|c}
    \toprule
    {Setting} & Soln. Acc. & Eqn. Acc. & Deriv. Acc.  \\
    \midrule
    \multicolumn{4}{c}{\dataname~+ Alg-514}         \\
    \midrule
    TE           & 32.5     & 31.5    & 29.5       \\
    TE$^*$            & 60.5     & 56.0    & 54.0       \\
    TD            & 62.0     & 53.0    & 59.5       \\
   \midrule
    TD - TE$^*$       & 1.5      & -3.0    & {\bf 5.5}  \\
    \midrule
    \multicolumn{4}{c}{\dataname~+ Dolphin}         \\
    \midrule
    TE           & 41.0     & 37.5    & 37.5       \\
    TE$^*$            & 58.5     & 55.5    & 51.5       \\
    TD            & 60.0     & 53.0    & 58.0       \\
   \midrule
    TD - TE$^*$       & 1.5      & -2.5    & {\bf 6.5}  \\
    \bottomrule
  \end{tabular}
  \caption{When combining two datasets, it is essential to reconcile templates across datasets. Here TE$^*$ denotes training on equations after reconciling the templates, while TE simply combines datasets naively. As TE$^*$ represents a more appropriate setting, we compare TE$^*$ and TD in this experiment.}
    \label{tab:combine}
\end{table}

Are solution and equation accuracy equally capable as derivation
accuracy at distinguishing between good and bad models?  To answer
this question, we train the solver under two settings such that one of
the settings has clear advantage over the other, and see if the
evaluation metrics reflect this advantage. The two settings are,

\paragraph{\sc \underline{TE} (Train on Equation)} Only the
$(\bx,\by)$ pairs are provided as supervision.  Similar to
\cite{Kushman14,Baidu2015:EMNLP}, the solver finds a derivation which agrees with the equation system and the solution, and trains on it. Note that the derivation found by the solver may be incorrect.

\paragraph{\sc \underline{TD} (Train on Derivation)} 
$(\bx,\bz)$ pairs obtained by the derivation annotation are used as
supervision. This setting trains the solver on human-labeled derivations.
Clearly, the TD setting is a more informative supervision
strategy than the TE setting.
TD provides the correct template and correct alignment (i.e. labeled derivation) as supervision and is expected to perform better than TE, which only provides the question-equation pair.

We first present the main results comparing different evaluation metrics on solvers trained using the two settings.
\subsection{Main Results}

We compare the evaluation metrics in
Table~\ref{tab:contributions}. We want to determine to what degree
each evaluation metric reflects the superiority of TD over TE.

We note that solution accuracy always exceeds derivation accuracy,
as a solver can sometimes get the right solutions even with the wrong
derivation. Also, solution accuracy is not as sensitive as derivation accuracy to improvements in the solver. For instance, solution accuracy
only changes by 2.4 on Dolphin-L when comparing TE and TD, whereas
derivation accuracy changes by 10.7 points. We found that the large gap on Dolphin-L was due to several alignment errors in the predicted derivations, which were detected by derivation accuracy. Recall that over 35\% of the problems in Dolphin-L have alignment ambiguities (Table~\ref{tab:comp}). In the TD setting, many of these errors made by our solver were corrected as the gold alignment was part of supervision.

Equation accuracy too has several limitations. For \dataname,
it cannot determine which solver is better and assigns
them the same score. Furthermore, it often (incorrectly) considers TD to be a worse
setting than TE, as evident from decrease in the scores (for instance, on {\sc Dolphin-L}).
Recall that equation accuracy attempts to approximate derivation
accuracy by choosing a random derivation agreeing with the equations,
which might be incorrect.

\paragraph{Study with Combining Datasets}
With several ongoing annotation efforts, it is a natural question to ask is whether we can leverage multiple datasets in training to generalize better. 
In Table \ref{tab:combine}, we combine \dataname's train split with other datasets, and test on \dataname's test split. \dataname's test split  was chosen as it is the largest test split with general algebra problems (recall Dolphin-L contains only number word problems). 

We found that in this setting, it was important to reconcile the templates {\em across} datasets. Indeed, when we simply combine the two datasets in the TE setting, we notice a sharp drop in performance (compared to Table~\ref{tab:contributions}).
However, if we reconciled all templates and then used the new equations for training (called TE$^*$ setting in Table ~\ref{tab:combine}), we were able to see improvements from training on more data. We suspect difference in annotation style led to several equivalent templates in the combined dataset, which got resolved in TE$^*$. Therefore, in Table \ref{tab:combine}, we compare TE$^*$ and TD settings.\footnote{In TE$^*$, the model still trains only using equations, without access to derivations. So TD is still better than TE$^*$.} 

In Table \ref{tab:combine}, a trend similar to Table \ref{tab:contributions} can be observed -- solution accuracy assigns a small improvement to TD over TE$^*$. Derivation accuracy clearly reflects the fact that TD is superior to TE$^*$, with a larger improvement compared to solution accuracy (eg., 5.5 vs 1.5). Equation accuracy, as before, considers TD to be worse than TE$^*$. 

Note that this experiment also shows that differences in annotation
styles across different algebra problem datasets can lead to poor
performance when combining these datasets naively. Our findings
suggest that derivation annotation and template reconciliation are
crucial for such multi-data supervision scenarios.

\subsection{Comparing Solvers}
\label{sec:sota}

To ensure that the results in the previous section were not an artifact of any limitations of our solver, we show here that our solver is competitive to other state-of-the-art solvers, and therefore it is reasonable to assume that similar results can be obtained with other automatic solvers.

In Table~\ref{fig:compare_kushman}, we compare our solver to {\sc Kazb}, the system of ~\newcite{Kushman14}, when trained
under the existing supervision paradigm, {\sc TE} (i.e., training
on equations) and evaluated using solution accuracy. We also report the best scores on each dataset, using {\sc Zdc} and {\sc Swllr} to denote the systems of~\newcite{Baidu2015:EMNLP} and~\newcite{MSRA2015}
respectively.  Note that our system and {\sc Kazb} are the only systems that can process all three datasets without significant modification, with our solver being clearly superior to {\sc Kazb}.

\begin{table}[t]
  \renewcommand*{\thefootnote}{\fnsymbol{footnote}}
  \centering
  \begin{tabular}{lccc}
    \toprule
    Dataset       & Ours & {\sc Kazb} & Best Result                                                                                              \\
    \midrule
    {\sc Alg-514} & 76.2 & 68.7       & {\bf 79.7} ({\sc Zdc})                                                                                   \\
    {\sc Dolphin-L} & 55.1 & 37.5       & 46.3\footnote[3]{
{\sc Swllr} also had a solver which achieves 68.0, using over 9000
semi-automatically generated rules tailored to number word problems.
We compare to their similarity based solver instead, which does not
use any such rules, given that the rule-based system cannot be applied to general word problems.}
    ({\sc Swllr}) \\ {\dataname} & 52.0 & 43.2 & -- \\ \bottomrule
  \end{tabular}
  \caption{Comparison of our solver and other state-of-the-art
    systems, when trained under {\sc TE} setting. All
    numbers are solution accuracy. See footnote for details on
    the comparison to {\sc Swllr}.}
  \label{fig:compare_kushman}
  \renewcommand*{\thefootnote}{\arabic{footnote}}
\end{table}
\subsection{Case Study}
\label{sec:case-study}
We discuss some interesting examples from the datasets, to show the limitations of existing metrics, which derivation accuracy overcomes.

\paragraph{Correct Solution, Incorrect Equation} In the following example from the {\sc Dolphin-L} dataset, by choosing the correct template and the wrong alignments, the solver arrived at the correct solutions, and gets rewarded by solution accuracy.
\begin{quotation}
\begin{small}
The sum of 2($q_1$) numbers is 25($q_2$). 12($q_3$) less than 4($q_4$) times one($q_5$) of the numbers is 16($q_6$) more than twice($q_7$) the other number. Find the numbers.
\end{small}
\end{quotation}
Note that there are seven textual numbers ($q_1,\ldots,q_7$) in the word problem. 
We can arrive at the correct equations (\{$m+n=25$, $4m-2n=16+12$\}), by the correct derivation,
\begin{equation*}
\begin{array}{cc}
    m + n = q_2 & q_4 m - q_7 n = q_6 + q_3.
\end{array}
\end{equation*}
However, the solver found the following derivation, which produces the incorrect equations (\{$m+n=25$, $2m-n=2+12$\}),
\begin{equation*}
  \begin{array}{cc}
    m + n = q_2 & \mathbf{q_1} m - \mathbf{q_5} n = \mathbf{q_7} + q_3.
  \end{array}
\end{equation*}
Both the equations have the same solutions ($m=13,n=12$), but the second derivation is clearly using incorrect reasoning.
\paragraph{Correct Equation, Incorrect Alignment} In such cases, the
solver gets the right equation system, but derived it using wrong
alignment. Solution accuracy still rewards the solver. Consider the
problem from the {\sc Dolphin-L} dataset, 
\begin{quotation}
\begin{small}
  The larger of two($q_1$) numbers is 2($q_2$) more than 4($q_3$) times the smaller. Their sum is 67($q_4$). Find the numbers.
  \end{small}
\end{quotation}
The correct derivation for this problem is,
\begin{equation*}
  \begin{array}{cc}
    m - q_3 n = q_2 & m + n = q_4.
\end{array}
\end{equation*}
However, our system generated the following derivation, which although results in the exact same equation system (and thus same solutions), is clearly incorrect due incorrect choice of "two",
\begin{equation*}
  \begin{array}{cc}
    m - q_3 n = \mathbf{q_1} & m + n = q_4.
\end{array}
\end{equation*}
Note that derivation accuracy will penalize the solver, as the alignment is not equivalent to the reference alignment ($q_1$ and $q_2$ are not semantically equivalent textual numbers).

\paragraph{Bad Approx.~in Equation Accuracy}
The following word problem is from the {\sc Alg-514} dataset:
\begin{quote}
\begin{small}
  \noindent Mrs. Martin bought 3($q_1$) cups of coffee and 2($q_2$) bagels and spent 12.75($q_3$) dollars. Mr. Martin bought 2($q_4$) cups of coffee and 5($q_5$) bagels and spent 14.00($q_6$) dollars. Find the cost of one($q_7$) cup of coffee and that of one($q_8$)~bagel.
\end{small}
\end{quote}
The correct derivation is,
\begin{equation*}
  \begin{array}{cc}
    q_1 m + q_2 n = q_3 & q_4 m + q_5 n = q_6.
\end{array}
\end{equation*}
However, we found that equation accuracy used the following incorrect
derivation for evaluation,
\begin{equation*}
  \begin{array}{cc}
    q_1 m + \mathbf{q_2} n = q_3 & \mathbf{q_2} m + q_5 n = q_6.
\end{array}
\end{equation*}
Note while this derivation does generate the correct equation system
and solutions, the derivation utilizes the wrong numbers and
misunderstood the word problem. This example demonstrates the needs to
evaluate the quality of the word problem solvers using the annotated
derivations.


\section{Related Work}
\label{sec:related}
We discuss several aspects of previous work in the literature, and how it relates to our study.
\paragraph{Existing Solvers}
Current solvers for this task can be divided into two broad categories
based on their inference approach -- {\em template-first} and {\em
  bottom-up}.  Template-first approaches like
~\cite{Kushman14,Baidu2015:EMNLP} infer the derivation $\bz=(T,A)$
{\em sequentially}. They first predict the template $T$ and then
predict alignments $A$ from textual numbers to coefficients.  In
contrast, bottom-up approaches~\cite{Verb14,MSRA2015,TACL692} {\em
  jointly} infer the derivation $\bz=(T,A)$. Inference proceeds by
identifying parts of the template (eg. ${\tt A}m+{\tt B}n$) and aligning
numbers to it ($\{\text{2} \rightarrow {\tt A}$,
$\text{3} \rightarrow {\tt B}\}$). At any intermediate state during
inference, we have a partial derivation, describing a
fragment of the final equation system (${\tt 2m +3n}$). While our experiments used a solver employing the template-first approach, it is evident
that performing inference in all such solvers requires constructing a
derivation $\bz=(T,A)$. Therefore, annotated derivations will be
useful for evaluating {\em all} such solvers, and may also aid in
debugging errors.

Other reconciliation procedures are also discussed (though
briefly) in earlier work. \newcite{Kushman14} reconciled templates by using a symbolic solver and removing
pairs with the same canonicalized form. \newcite{Baidu2015:EMNLP}
also reconciled templates, but do not describe how it was
performed. We showed that reconciliation is important for correct evaluation, for reporting dataset complexity, and also when combining multiple datasets.

\paragraph{Labeling Semantic Parses}
Similar to our work, efforts have been made to annotate semantic
parses for other tasks, although primarily for providing supervision.
Prior to the works of \newcite{liang-jordan-klein:2009:ACLIJCNLP} and
\newcite{clarke2010driving}, semantic parsers were trained using
annotated logical forms~\cite[\em inter
alia]{ZelleMo96,ZettlemoyerC05,wong2007learning}, which were expensive
to annotate.  Recently, \newcite{YRMCS16} showed that labeled semantic
parses for the knowledge based question answering task can be obtained
at a cost comparable to obtaining answers. They showed significant
improvements in performance of a question-answering system using the
labeled parses instead of answers for training. More recently, by treating
word problems as a semantic parsing task, \newcite{UCCY16} found that
joint learning using both explicit (derivation as labeled semantic
parses) and implicit supervision signals (solution as responses) can significantly
outperform models trained using only one type of supervision signal.
%
%

\paragraph{Other Semantic Parsing Tasks}
We demonstrated that response-based evaluation, which is quite popular
for most semantic parsing problems~\cite[{\em inter
  alia}]{ZelleMo96,BCFL13,liang2013learning} can overlook reasoning
errors for algebra problems. A reason for this is that in algebra word
problems there can be several semantic parses (i.e., derivations, both
correct and incorrect) that can lead to the correct solution using the
input (i.e., textual number in word problem). This is not the case for
semantic parsing problems like knowledge based question answering, as
correct semantic parse can often be identified given the question and
the answer. For instance, paths in the knowledge base (KB), that
connect the answer and the entities in the question can be interpreted
as legitimate semantic parses. The KB therefore acts as a constraint
which helps prune out possible semantic parses, given only the problem
and the answer. However, such KB-based constraints are unavailable for
algebra word problems.

\section{Conclusion and Discussion}
We proposed an algorithm for evaluating derivations for word problems. We also showed how derivation annotations can be easily obtained by only involving annotators for ambiguous cases.
We augmented several existing benchmarks with derivation
annotations to facilitate future comparisons.
Our experiments with multiple datasets also provided insights into the right approach to combine datasets -- a natural step in future work. Our main finding indicates that derivation accuracy leads to a more accurate assessment of
algebra word problem solvers, finding errors which other metrics overlook. While we should strive to build such solvers using as little
supervision as possible for training, having high quality annotated
data is essential for correct evaluation.

The value of such annotations for evaluation becomes more immediate
for online education scenarios, where such
word solvers are likely to be used. Indeed, in these cases, merely arriving at the correct solution, by
using incorrect reasoning may prove detrimental for teaching purposes. We believe derivation based evaluation closely mirrors how humans are evaluated in schools (by forcing solvers to show ``their work'').

Our datasets with the derivation annotations have applications beyond accurate evaluation. For instance, certain
solvers, like the one in ~\cite{roy2015solving}, train a {\em relevance classifier} to
identify which textual numbers are relevant to solving the word problem. As
we only annotate relevant numbers in our annotations, our
datasets can provide high quality supervision for such classifiers. The datasets can also be used in evaluation test-beds, like the one proposed in \cite{MAWPS}.

We hope our datasets will open new possibilities for the community to
simulate new ideas and applications for automatic problem solvers.


\section*{Acknowledgments}
The first author was supported on a grant sponsored by DARPA under agreement number FA8750-13-2-0008.
We would also like to thank Subhro Roy, Stephen Mayhew and Christos Christodoulopoulos for useful discussions and comments on earlier versions of the paper. 

\bibliography{algebra_long,cited}
\bibliographystyle{eacl2017}

\appendix

\section{Creating~\dataname}
We crawl over 100k problems from \url{http://algebra.com}. The 100k
word problems include some problems which require solving non-linear
equations (e.g. finding roots of quadratic equations). 
We filter out these problems using keyword
matching. We also filter problems whose explanation do not contain a
variable named ``x''. This leaves us with 12k word problems.

\paragraph{Extracting Equations}
A word problem on \url{algebra.com} is accompanied by a detailed explanation
provided by instructors. 
In our crawler, we use simple pattern matching rules to extract all
the equations in the explanation. 
The problems often have sentences which are irrelevant to solving the
word problem (e.g. ``Please help me, I am stuck.'').  During cleaning,
the annotator removes such sentences from the final word problem and
performs some minor editing if necessary.\footnote{In some cases, some
  of the numbers in the text are rephrased (``10ml'' to ``10 ml'') in
  order to allow NLP pipeline work properly.}

1000 problems were randomly chosen from these pool of 12k problems, which were then shown to annotators as described earlier to get the derivation annotations. 
\section{Proof of Correctness (Sketch)}
For simplicity, we will assume that EquivTNum is empty. The proof can easily be extended to handle the more general situation. 

\paragraph{Lemma 1.} The procedure {\sc TemplEquiv} returns $\Gamma \neq \emptyset$ iff templates $T_1$, $T_2$ are equivalent (w.h.p.).
\paragraph{Proof}
First we prove that with high probability we are correct in claiming that a $\gamma$ found by the algorithm leads to equivalence.
Let probability of getting the same solution even when the template are not equivalent be $\epsilon(T_1,T_2,\gamma) < 1$. The probability that solution is same for R rounds for $T_1, T_2$ which are not equivalent is $\le \epsilon^R$, which can be made arbitrarily small by choosing large R. Therefore, with a large enough R, obtaining $\Gamma \neq \emptyset$ from {\sc TemplEquiv} implies there is a $\gamma$ under which templates generate equations with the same solution, and by definition, are equivalent.

Conversely, if templates are equivalent, it implies $\exists \gamma^*$ such that under that mapping for any assignment, the generated equations have the same solution. As we iterate over all possible 1-1 mappings $\gamma$ between the two templates, we will find $\gamma^*$ eventually.

\paragraph{Proposition}
Algorithm~\ref{algo:eval} returning 1 implies derivations $(T_p,A_p)$ and $(T_g,A_g)$ are equivalent.
\paragraph{Proof} Algorithm returns 1 only if {\sc TemplEquiv} found a $\Gamma \neq \emptyset$, and $\exists\gamma \in \Gamma$, following holds
$$
(q,c) \in A_g \iff (q,\gamma(c)) \in A_p
$$
i.e., the corresponding slots aligned to the same textual number.
{\sc TemplEquiv} found a $\Gamma \neq \emptyset$ implies templates are equivalent (w.h.p). Therefore, $\exists \gamma \in \Gamma$ such that the corresponding slots aligned to the same textual number implies the alignments are equivalent under mapping $\gamma$. Together they imply that the derivation was equivalent (w.h.p.).


\end{document}